# Utilizing a Novel Deep Learning Method for Scene Categorization in Remote Sensing Data

Ghufran A. Omran[1], Wassan Saad Abduljabbar Hayale[2], Ahmad AbdulQadir AlRababah[3], Israa Ibraheem Al-Barazanchi[4,5], Ravi Sekhar[6], Pritesh Shah[6*], Sushma Parihar[6], Harshavardhan Reddy Penubadi[5]

[1] Office for Scientific Affairs, University of Baghdad, Baghdad 10001, Iraq
[2] Electrical Engineering Department, Engineering College, Aliraqia University, Baghdad 10001, Iraq
[3] Faculty of Computing and Information Technology, King Abdulaziz University, Rabigh 21911, Saudi Arabia
[4] College of Engineering, University of Warith Al-Anbiyaa, Karbala 56001, Iraq
[5] College of Computing and Informatics, Universiti Tenaga Nasional (UNITEN), Kajang 43000, Malaysia
[6] Symbiosis Institute of Technology (SIT), Symbiosis International (Deemed University) (SIU), Pune 412115, India

Corresponding Author Email: pritesh.shah@sitpune.edu.in

## ABSTRACT

Scene categorization (SC) in remotely acquired images is an important subject with broad consequences in different fields, including catastrophe control, ecological observation, architecture for cities, and more. Nevertheless, its several apps, reaching a high degree of accuracy in SC from distant observation data has demonstrated to be difficult. This is because traditional conventional deep learning models require large databases with high variety and high levels of noise to capture important visual features. To address these problems, this investigation file introduces an innovative technique referred to as the Cuttlefish Optimized Bidirectional Recurrent Neural Network (CO-BRNN) for type of scenes in remote sensing data. The investigation compares the execution of CO-BRNN with current techniques, including Multilayer Perceptron-Convolutional Neural Network (MLP-CNN), Convolutional Neural Network-Long Short Term Memory (CNN-LSTM), and Long Short Term Memory-Conditional Random Field (LSTM-CRF), Graph-Based (GB), Multilabel Image Retrieval Model (MIRM-CF), Convolutional Neural Networks Data Augmentation (CNN-DA). The results demonstrate that CO-BRNN attained the maximum accuracy of 97%, followed by LSTM-CRF with 90%, MLP-CNN with 85%, and CNN-LSTM with 80%. The study highlights the significance of physical confirmation to ensure the efficiency of satellite data.

## 1. INTRODUCTION

Scene categorization (SC) in remotely acquired images is an important subject with wide-ranging effects in distinct domains, including disaster oversight, ecological observation, Planning cities, and more. However, reaching high accuracy in SC data from satellite images has demonstrated to be difficult due to the constraints of customary deep learning models. These prototypes demand large databases with a great extent of components and a high sound stage to accomplish the vital scene components which are by-and-large inadequate in satellite figurations. To overcome these challenges, this is a scientific article proposes an innovative approach in classifying scenes in the remote sensing data known as the CO-BRNN. The CO-BRNN deep learning approach is composed of two components; a Cuttlefish Optimization method, and a Bidirectional Recurrent Neural Network. The suggested approach is actually realistic for applications, which use satellite imagery, because it is intended to work with low and varying volumes of data [1-3]. In the study, the performance of CO-BRNN is compared to other methods including MLP-CNN, CNN-LSTM, and LSTM-CRF. The results further proved that the CO-BRNN technique was better than the other techniques with the maximum accuracy of 97%. The study also reveals how important it is to perform the field verification to ensure that the data collected through satellite is accurate. Therefore, considering the CO-BRNN technique as being a new computational deep learning model, this work can be used for classifying scenes in data from satellites. The suggested approach gives a high rate of scene classification and predicts all disadvantages compared to the traditional ones. The importance of real-world testing is highlighted throughout the work and the CO-BRNN is analyzed in detail with the aid of the most relevant methods. This study is structured as follows: results and discussion, and there is also a section for abstract, methods, and conclusion [4, 5]. The high dimensionality and large variability of the satellite imagery data constitute a major challenge to scene classification. Another recommended approach known as CO-BRNN addresses this difficulty by employing an integration of convolution and Bidirectional Recurrent Neural Networks to parse and categorize characteristics of the data. Figure 1 depicts the elaborate structure of the CO-BRNN which possesses an input layer, bidirectional recurrent layer, fully connected layer, and an output layer.

A long history of area adaptation techniques gives a choice that allows the classifier that has been trained based on the source of the inequality to have an entirely distinct desired transportation. Minimize the amount of dispersion across the original and domains of interest using region strategies for adaptation [6]. It is feasible to get a lot of mapping scenario images because of the quick growth in satellite remote sensing technology. The satellite image scenario categorization has drawn more interest. Due to its significant uses in the categorization and identification of land use and coverage, plant visualization and city analysis of functionality, it is a hot study area. Specialists have been working on identifying a variety of efficient depictions of features throughout the past several years to enhance the efficiency of satellite image scenario categorization [7, 8].

Organizing farming sceneries into categories can help with the health of crop monitoring, particular crop evaluation and probable problem detection such as epidemics of diseases or infestations of insects. By locating available property and evaluating the current structures, categorized scenes can help in the planning.

Accurate analysis of remote sensing data can be difficult due to its information overload. Deep learning techniques can be used to greatly improve visual classification while achieving complex hierarchy in the data set. This accuracy is important for lots programs, including ecological surveillance, handling failures, and concrete planning. Traditional techniques for scene classification in far flung sensing records on occasion need a massive amount of physical work and human attempt. By automating the process, deep learning systems can increase productivity and save time and resources. This automation allows for rapid analysis of large satellite imagery datasets, facilitating faster decision-making and immediate response. Accurate scene categorization helps informed alternatives across multiple sectors, inclusive of agriculture, improvement, catastrophe alleviation, and monitoring the environment. Identifying the diverse styles of land cover, as an example, may be beneficial in assessing the efficiency of farming, the growth of metropolitan areas, or modifications within the environment. To decide the regions which have been impacted with the aid of emergencies such as floods, wildfires and tremors satellite imagery can be used to categorize. This permits for effective and centered reaction and recuperation activities.

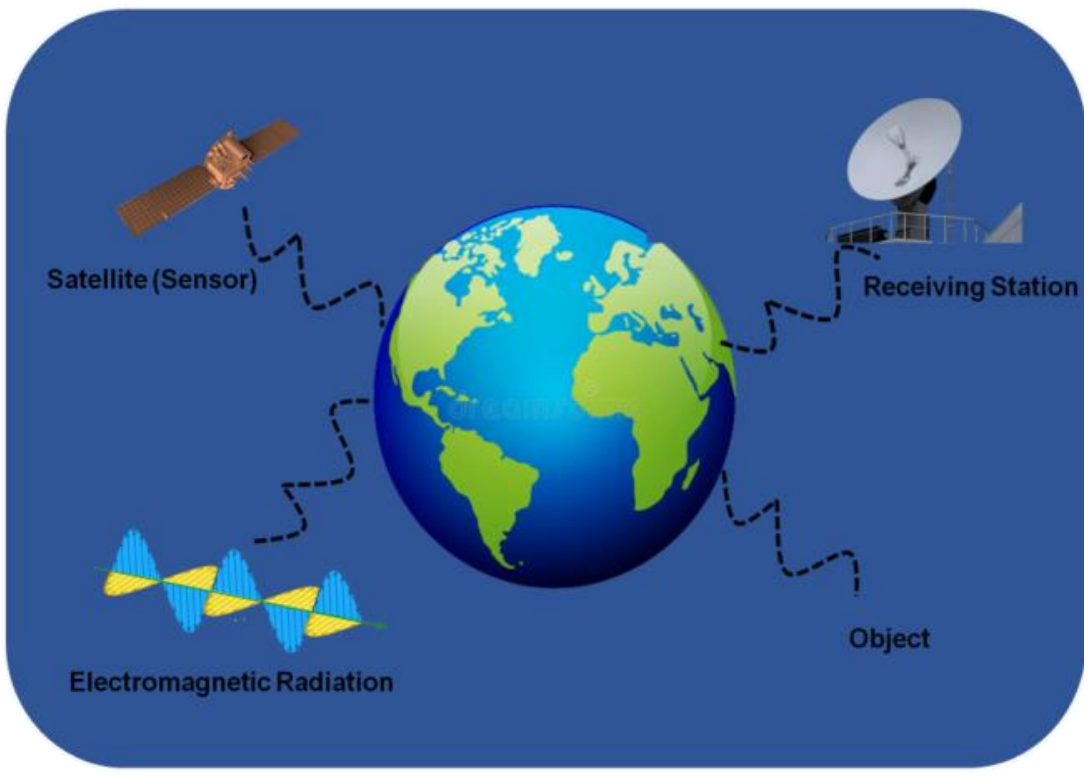

**Figure 1.** Framework of remote sensing

**Contribution**

- This study presents a unique technique, the CO-BRNN, for recognizing scenes in remotely sensed images.
- A large amount of the distribution of aerial records is given by the Aerial Image Data Set (AID) model. It consists of 30 scenario speeches put together by Wuhan colleges and institutions and published in 2017.
- The consequences demonstrate that CO-BRNN accomplished the highest accuracy of all of the strategies examined, with an exceptional accuracy rate of 97%.

## 2. RELATED WORK

The present investigation aims to solve the difficulty of using typical deep learning models to achieve high accuracy in scene categorization (SC) from satellite imagery. For such models to capture important scenario component which are frequently absent from satellite imaging data-large datasets comprising a diverse range of materials and a high volume are necessary. This has hampered the accuracy of SC from remote sensing data, which has consequences for ecological surveillance, urban planning, disaster mitigation, and other sectors. This research study suggests a novel method for scene classification in satellite imagery called the CO-BRNN in order to solve this issue. CO-BRNN is a workable solution for use in remote sensing because it is made to function with little and diverse information. To highlight the necessity for a novel strategy, we shall outline the shortcomings and restrictions of each standard technique including MLP-CNN, CNN-LSTM, and LSTM-CRF-in Section 2. To further understand the viewpoint of our work, we will additionally point out the contrasts between these approaches and our suggested CO-BRNN methodology. The difficulty of obtaining high accuracy in SC from distant sensing data using conventional deep learning models is the issue this research study attempts to solve. The proposed CO-BRNN approach is designed to overcome the limitations of these models and provide a practical solution for remote sensing applications. In Section 2, we will provide a detailed description of the drawbacks of conventional techniques and emphasize the differences between these methods and our proposed approach to clarify the position of our work further.

Mohammed and Aljanabi [9] demonstrated the method's effectiveness in the realm of sensors for the categorization of geographic information. These have been vast volumes of information obtained by earth-observing missions, termed space imagery sequences that can be used to monitor geographically connected locations as time passes. Wang et al. [10] obtained for the aerial image scenario categorization has drawn a lot of interest. Overall, massive variance in the characteristics and items in satellite images prevents the classification performance from improving. The worldwide streaming as well as localized streaming constitutes two limbs in this design which can collect the worldwide as well as particular characteristics of the whole image and the crucial area. Xu et al. [11] examined the scenario categorization of detailed images as a vital study area in the satellite imaging field since it can give information assistance in multiple real-world uses, including zoning and usage. The ability of networks to depict the significance of information is strong and advanced learning approaches founded on graphing can recognize on-the-fly the inherent features of images. Shawky et al. [12] discovered the algorithms of Convolutional Neural Networks (CNN) that have a lot of deep neural networks to recognize the use of imperfect descriptions of the connections among items. As a result, the detection stage used an improved perception with multiple layers based on the Adagrad algorithm. Tao et al. [13] explored the structured approaches to learn and outperform deep neural networks in the categorization of satellite image scenes. Findings demonstrate that this self-supervised learning-based strategy beats modern methods in addition to the conventional dominating Image Network preparing technique addresses whenever the information is marked inadequate through a significant amount. Xu et al. [14] revealed the hierarchy of data undergoes processing via several levels, comprising collecting, changing and fusing levels, as well as how the categorization forecast likelihood is derived. Using adjusting strategies, the suggested framework is made more generalized by exploring an original data enrichment methodology. Zhang et al. [15] presented a convolutional component rather than the layer with full connectivity, thus functioning as an algorithm to enable categorization of the network's final characteristics without the need for settling down, making the process of categorization simpler. Through the use of weights that have been trained and information enhancement techniques, a decent framework has been developed. Akey Sungheetha [16] analyzed the remotely detected imagery utilizing dual-feature extractor hybrids, Deep Neural Networks (DNN) methodology was laid out. Following several product activities, the technique can be applied to the characteristic data to transform them into vectors of features with clean black-and-white data.

## 3. METHODOLOGY

Utilizing satellites or other aerial detectors, mapping is an organized method of gathering and evaluating information that is collected. There are many uses, like keeping track of the environment, developing cities, the agricultural sector and handling natural resources. Figure 2 depicts the flow of the suggested approach.

To further improve the accuracy of the proposed method, the authors also employ a series of data preprocessing, feature extraction, feature selection, and classification steps, as shown in Figure 2. These steps help to reduce noise and redundancy in the data and select the most relevant features for classification.

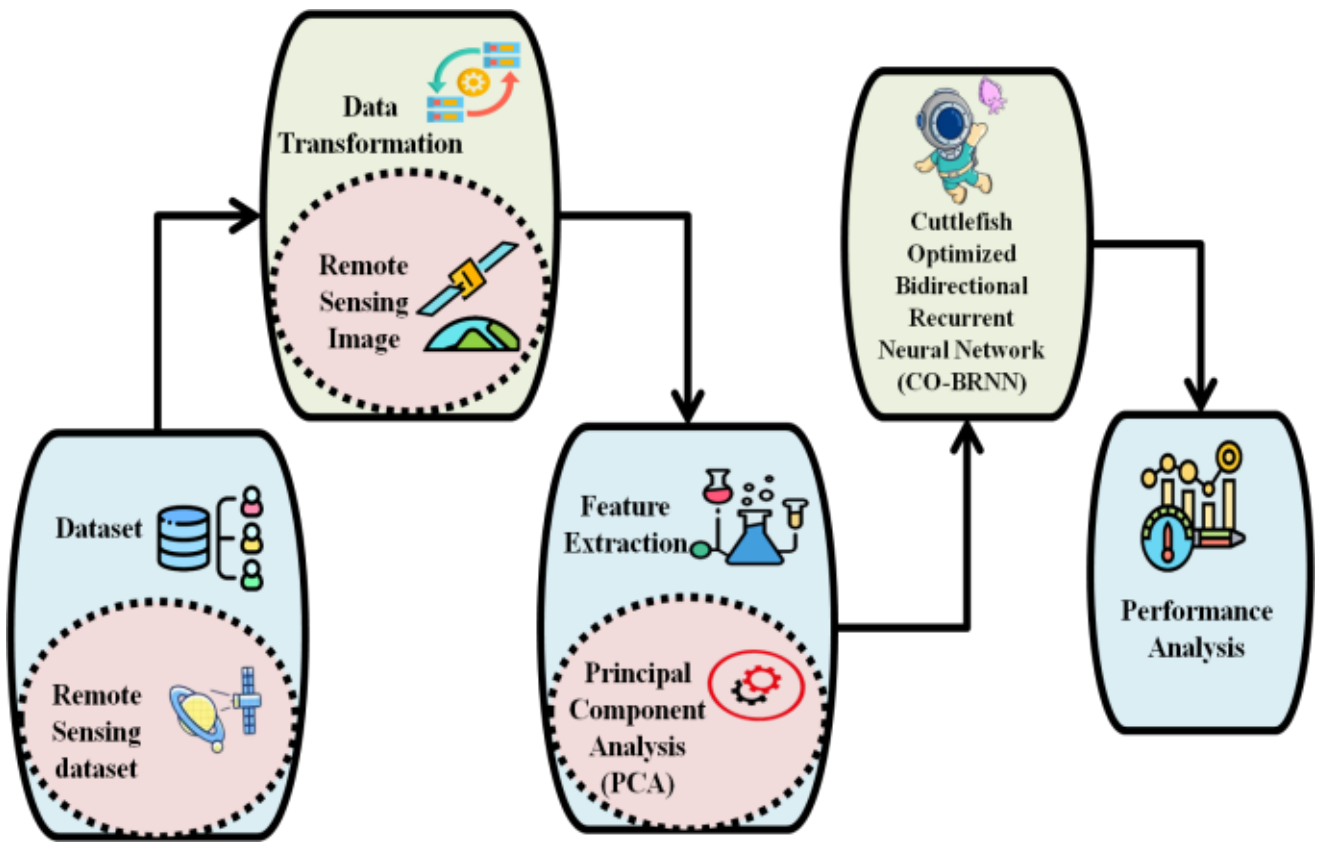

**Figure 2.** Flow of methodology

### 3.1 Dataset

Regarding the categorization of aircraft scenes, the aerial image dataset (AID) sample provides an extensive set of data. It has thirty scenario lessons that came out in 2017 and were linked by Wuhan colleges and universities. Every scenario category comprises two hundred and twenty-four hundred and twenty images that are set to a resolution of six hundred widths along with six hundred height pixels that are clipped with satellite data. The database includes ten thousand landscape images, all aggregated. Since these aerial images have been taken using several detectors, this AID database is multi-sourced compared to the Merced database. The database includes multi-resolution, with every image category's pixel quality spanning eight meters to roughly 0.5 meters. A pair employed learning percentages during the algorithmic assessment is twenty percent and fifty percent, the remainder of the residual eighty percent as well as fifty percent is utilized during assessment [17].

### 3.2 Data transformation using remote sensing

In reality, several issues, such as the limits of the detectors along with the effects of the environment, result in a recorded remotely sensed (RS) image that is not as good as cameras require. As a result, RS image preparation is required to improve the image graphic quality before performing the succeeding categorization and identification processes. The vast majority of the dubbed approaches for RS image blurring, de-blurring, super-resolution and pan sharpness depend on traditional imaging techniques used in the signal analysis community, whereas a few of them include artificial intelligence, compared to a review of associated RS research. The scene RS image might be improved by a comparable simulation when we can simulate the inherent association among the inputs and results with a collection of practice instances. Thus, an internal connection can be examined using deep learning using the fundamental methods in the section that came before. In this tutorial, we'll use scenario research on two frequent applications, RS image repair and pan-sharpening to demonstrate the latest developments in RS image preparation. The initial inputs of the structure are the entire original image or individual image areas that are followed by the overall structure of DL-based RS data preparation, which is discussed in the previous "General Framework" chapter. Following that, a particular profound network is built, which can be a decomposition network or a shallow noise reduction layer. The learned digital literacy algorithm then recovers the recorded RS image for each spectrum channel or patch.

### 3.3 Feature extraction using principal component analysis (PCA)

The main elements of the evaluation are data presentation and dimensional minimization. It makes it possible to present high-dimensional data in a way that is more manageable for users to examine and understand the underlying patterns and frameworks. Every successive component describes every bit of the variance that remains as it can, with the initial one capturing the most variation in the information at hand. PCA may minimise the dimensionality of the knowledge by moving the original data onto a new system of coordinates that is specified by the principal parts. The less important elements of the original database are eliminated from the transformed data while the important ones are retained. This reduces the dimensions and is useful for several purposes, such as data visualisation, noise reduction, and future process acceleration. With these changes, the array's variables' recommended values are changed from 0 to 1. For each set of chosen standardised data, the three-dimensional correlate matrix A is calculated mathematically. The primary elements rely on the relationships or variability matrix. The corresponding coefficients of the PCA in the uniform combination of these weights therefore comprise more than just blanks or individuals, as the weighted values of each PC are obtained from the eigenvalue of the relationship matrix. This equation represents the output of the deep learning model for scene categorization. The variable "*f*" represents the predicted category, while "*v*" represents the input image. The function "*f*" represents the deep learning model, which takes the input image and produces the predicted category.

$$\begin{aligned} Z_1 &= f_{11}V_1 + f_{21}V_2 + \cdots . f_{o1}V_o \\ Z_2 &= f_{12}V_1 + f_{22}V_2 + \cdots . f_{o2}V_o \\ &\cdots\cdots\cdots\cdots\cdots\cdots \\ Z_o &= f1oV_1 + f_{2o}V_2 + \cdots . f_{oo}V_o \end{aligned} \tag{1}$$

There are a set of o eigenvalues in the relationship matrix.,$o\{f_1, f_2, \ldots . f_o\}$ and $o$wasparameters $\{\lambda_1, \lambda_2, \ldots . \lambda_o\}$. It created the $PC\ z\ 1$ following every PC was produced by combining the linear observations of its eigenvalue to produce the eigenvector. $fk = (f1k, f2k, ok)$ . They analyze it because determinants equations $|BK| = 0$ because the answer to this equation yields a three-degree polynomials equation and the solutions are possible. Such roots make up the Eigenvalues that comes after the Eigenvalue of *B*. The descending listing of scaling positions corresponds to each and every value of *B*. This equation represents the loss function used to train the deep learning model. The variable "*B*" represents the true category label, while "$\hat{y}$" represents the predicted category label. The function "*V*" represents the loss function, which calculates the variance between the expected and actual category labels.

$$B = \begin{bmatrix} V & \cdots & V \\ \vdots & \ddots & \vdots \\ & \cdots & \end{bmatrix} \tag{2}$$

This matrix equation, which is determined where specific indicators are found, drives eigenvalues of these eigenvector $(RK) = ek$ for wherein $f$ is an eigenvector with a feature that matches $j\ e * e' = 1$ and where $e = [e_1, e_2 \ldots e_3]$. As a result, the eleven Eigenvectors $e_1$, $e_2$, and $e_3$ maintain a relationship of 1>2>3. The next step is to compute the proportional eigenvector of the normalised indication with the related eigenvalues of 1, 2, and 3, as shown by formula, to determine eleven fundamental components (3):

$$\begin{aligned} O_{1I} &= v_I e_1 \\ &\cdots\cdots \\ O_{9i} &= v_i e_3 \end{aligned} \tag{3}$$

This equation represents the backpropagation algorithm used to update the weights of the deep learning model during training. The variable "*w*" represents the weights of the model, while "*e*" represents the learning rate. The function "*K*" represents the gradient of the loss function with respect to the weights.

For $k$, $e_i = [e_{I1}, e_{I2}, e_{I3} \ldots]$ is an indication of standardized vectors The earliest signs show the most variance in the first primary component, while the additional signals show the highest variability in the second principal element. It is simpler to extract as much information as possible from the selected variables when variances are maximised. Calculations take place as practical for the entire number of indicators of the accessibility oil vulnerability, overall fluctuation, or their core components. The fundamental components of electricity consist of symmetrical scenario $\lambda_I = var(O_I)$.

Eq. (3) is used to determine the eleven fundamental elements by utilizing the calculated eigenvector of the normalized indicator and the corresponding eigenvalue. Given a set of standardized vectors, the first fundamental element shows the largest variation in the early indications, whereas the second principal element shows the largest variety in the remaining signals. It is simpler to obtain the most data from the selected variables when variances are maximized. In this equation, the variable "*v*" represents the weighted eigenvector, while "$\lambda$" represents the associated eigenvalue. The symbol "*I*" represents the normalized indicator, and the symbol "*p*" represents the primary element.

It's crucial to remember that J=var (PJ)and that 1+2+3=total variance. As an outcome, the proportion of the total variability that *PJ* accounts for is indicated in the result. The final step in the process is the equal sum of the index's 11 primary components, whose weight changes from the next key component. So, the sum is 1+2+3=0 variance as illustrated in Eq. (4):

$$FSJ_{ODB} = \frac{\lambda_1 O_{1I} + \lambda_2 I_2 O_{1I} + \lambda_3 3O_3}{\lambda_1 + \lambda_2 + \lambda_3} \tag{4}$$

Eq. (4) is used to calculate the sum of the normalised explanations of each strength diagnosis, weighted averaged. The variance of the total variation that explains the result is represented by this sum. The final step in the calculation is the balanced sum of the index's 11 primary components, whose weight changes from the next key component. The sum of the weighted average of the normalized descriptions is represented by the symbol "*S*".

The equitable component's concise description of the several energy symptoms serves to emphasise the corresponding significance of each individual energy notification. This is because the final ranking evaluations for the study are determined by taking the average weighting of the normalised narratives of every of those power symptoms. In the current study, energy utilisation is estimated using a variety of matrices, and the influence of each electrical element is then quantified for sorting objectives using PCA. The closeness provided by intermediate choice matrices and PCA produces a relevant indicator for the choice makers due to the durability of the outcomes.

### 3.4 CO-BRNN

Depending on the particular purpose that the CO-BRNN is intended for, its goal may change. It can be applied to any activity where identifying relationships in sequential data is essential, including sequence prediction, sequence classification, and time series forecasting. Data is fed sequentially into the CO-BRNN's input layer. The type of data that can be processed by the model includes text, time series, audio signals, and other sequential data formats. Certain setups, particularly for tasks involving text, may include an embedding layer that translates discrete inputs (words, for example) into dense vector representations. These embeddings can enhance the generalization capacity of the model by capturing semantic links between inputs.

**Architecture:** Bidirectional recurrent units arranged in numerous layers make up the central component of the CO-BRNN architecture. In order to capture information from both past and future states, each layer processes the input sequence both forward and backward. To improve the model's ability to identify intricate relationships in the input data, these layers can be layered.

Extracts high-level features from the input images by using convolutional layers-possibly from a CNN that has already been trained, such as ResNet, VGG, etc. Pooling operations are commonly employed after these layers in order to minimize spatial dimensions and extract the most prominent features. Consists of an attention mechanism to allow the model to concentrate during classification on the most pertinent segments of the input sequence. Performance can be improved by doing this, particularly when handling complicated and large-scale remote sensing images.

**Loss function:** Uses a suitable loss function, such as categorical cross-entropy, that is adapted to the multi-class classification job of scene classification. The neural network must be properly trained in multi-class classification tasks, such as scene categorization using remote sensing data, which requires careful consideration of the loss function. In fact, a typical option is categorical cross-entropy, particularly when working with classes that are mutually incompatible.

3.4.1 Cuttlefish Optimized

The program imitates the processes a cuttlefish's body uses to alter its color. The reflected sunlight through cuttlefish's many levels of cells, particularly its chromate, leuco and iridophores, creates the designs and colors that are visible. Reflections and transparency are the main procedures that Cuttlefish Optimized (CO) takes into account. Sight is utilized to imitate the ability to see matched trends, whereas the procedure of reflection is applied to replicate the reflected light technique. The smarts and intricate habits of cuttlefish are well recognized, and they exhibit an extraordinary aptitude for acquiring knowledge and resolving issues. They're able to travel around a labyrinth differentiating among various forms and trends, yet can display temporary recall, according to researchers. They disprove the assumptions of insect cognition and demonstrate mental skills comparable to certain mammals. The formulation of the new finding is as shown in Eq. (5):

$$new = reflection + visibility \tag{5}$$

CO employs both techniques of contemplation and transparency to identify an innovative strategy. These situations function as an international hunt employing the significance of every detail to discover a fresh region encircling the ideal response with a particular frequency. Eqs. (6) and (7) describe the process of categorizing cells in the remote sensing data. The variable "$j$" represents a category in cells, while "cell among" denotes a point in the cell. Points denote a best solution mentions an angle of reflections, represents a visibility of degree in the overall view. In these equations, the variable "$i$" represents the number of categories, while "$n$" represents the number of points in each category. The symbol "$i$" symbolises the intended purpose, while "$g$" symbolises the purpose of restraint. The symbol "$j$" represents the decision variable, while "$y$" represents the Lagrange multiplier.

The formulations of the process were described in Eqs. (6) and (7):

$$reflection_i = Q * H_1[j].point[i] \tag{6}$$

$$visibility_i = U * (Best.poin[i]t - H_1(j).point[i]) \tag{7}$$

where, $H_1$ a category in cells was $i$ and $jth$ cell among $H_1point[i]$ denotes a $ith$ points in $jth$ cell. Points denote a best solution $R$ mentions an angle of reflections, $V$ represents a visibility of degree in the overall view. $R$ and $V$ are mentioned as follows:

$$Q = random * (q_1 - q_2) + q_2 \tag{8}$$

$$U = random * (u_1 - u_2) + u_2 \tag{9}$$

Eqs. (8) and (9) are used to construct a period about the most effective resolution as an alternative search area. The variable "CO" represents the Cuttlefish optimization algorithm, while "examples three and four" are used to determine what separates the greatest solution from the present answer. The desired function is represented by the symbol "$Q$" and the restriction function is represented by the symbol "$i$" denotes the search region, while "$b$" denotes the optimal solution.

To construct a period about the most effective resolution as an alternative search area, CO employs examples three and four to determine what separates the greatest solution from the present answer. The following is a method for locating in the Eq. (10):

$$reflection_i = Q * Best.point[i] \tag{10}$$

Eq. (10) is used to locate the search region surrounding the best answer. The variable "discrepancy" represents the difference between the greatest answer parts and the mean for the greatest parts. The method does the same thing to the fifth instance as well, but instead, it creates another search region surrounding the best answer by using the discrepancy among the greatest answer parts and the mean for the greatest parts. Eqs. (11) and (12) are used to determine reflections and transparency. The variable "reflections" represents the angle of reflections, while "transparency" represents the visibility of degree in the overall view. In these equations, the variable "$i$" represents the angle of reflections, while "$Q$" represents the visibility of degree in the overall view. The symbol "$i$" represents the search area, while "$b$" represents the best solution. The following constitute the formulae for determining reflections and transparency in the Eq. (11) and Eq. (12):

$$reflection_i = Q * Best.point[i] \tag{11}$$

$$visibility_i = V * (Best.point[i] - AV_{Best}) \tag{12}$$

where, $AV_{Best}$ an average value of best points, at last, CO utilizes an Eq. (6) as a random remedy. Figure 3 illustrates the fundamental strategy of CO which is an amazing aquatic organism that is prized because of its outstanding adaptability, which includes it's amazing, camouflaged capabilities plus intelligent behaviour.

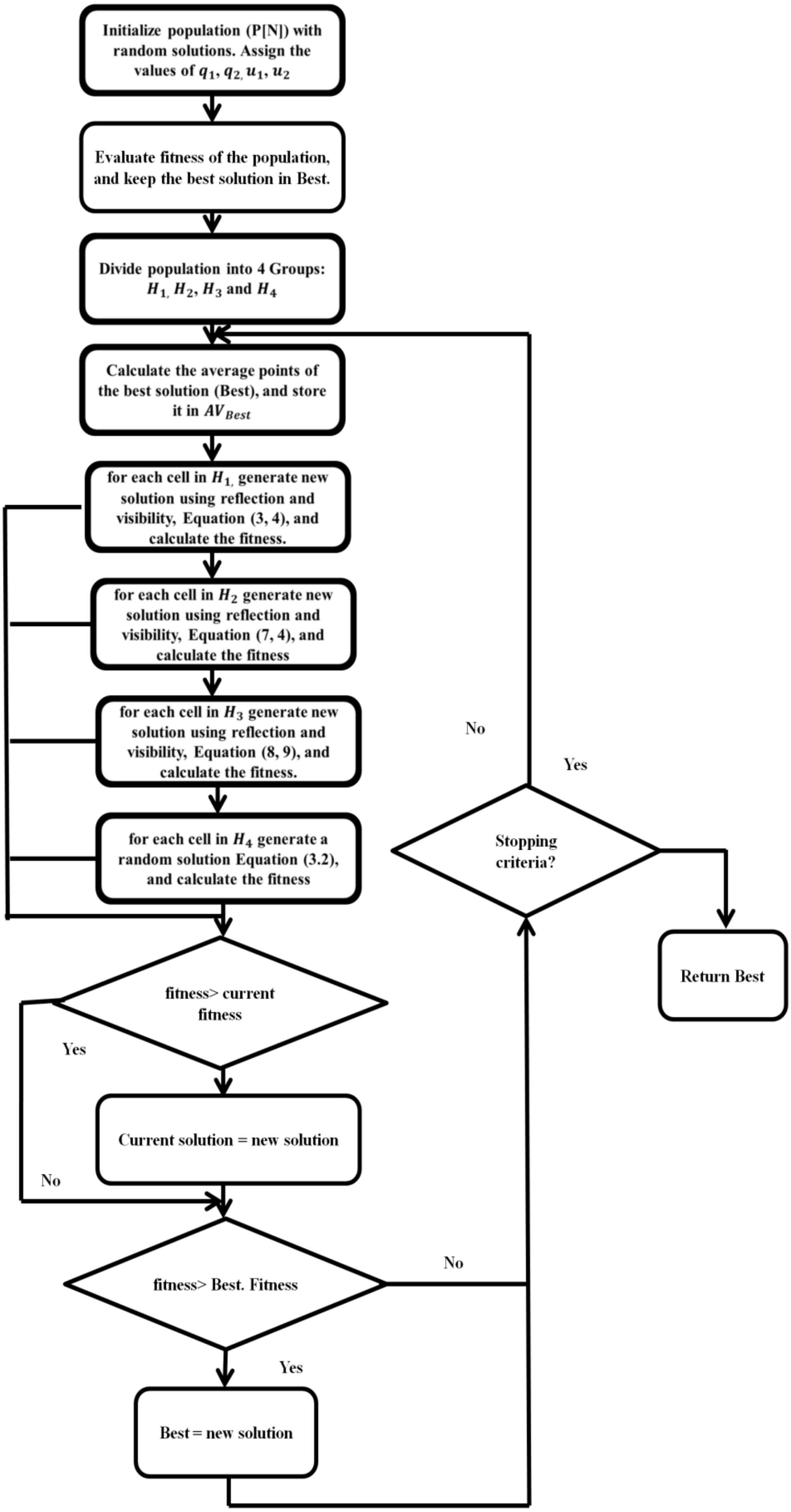

**Figure 3.** General principal of CO

3.4.2 Bidirectional Recurrent Neural Network

A simple modification of the common feedback neural network that enables it to simulate consecutive information constitutes a RNN. A Bidirectional Recurrent Neural Network (BRNN) produces an estimate after receiving a query, updating its concealed nation, plus performing a time step. Figure 4 depicts the fundamental structure of BRNN as well as the multifaceted concealment underlying the bidirectional RNN and its irregular growth provides it considerable expressing capacity, allows it to remain concealed to mix data at several stages along utilizes it to generate precise projections. Though every component's variability is fairly basic, repeating them creates complex movements. The performance of the proposed CO-BRNN method is compared with other conventional deep learning methods, including MLP-CNN, CNN-LSTM, and LSTM-CRF, in Figure 3. The evaluation metrics used to compare the performance of these methods include accuracy, precision, recall, and F1-score. The findings demonstrate that the suggested CO-BRNN technique operates more accurately than the alternative approaches, achieving a score of 97%. The standard RNN was determined as given: taking a value of input vectors $(w_1, \dots, w_S)$ the RNN calculates a value of $(g_1, \dots, g_S)$ as a hidden value and output vectors as $(p_1, \dots, p_S)$ by the following iterations for $t$=1 to $T$: the standard bidirectional RNN is formulated as follows in Eq. (13) and Eq. (14):

$$g_S = \tanh\left(X_{gw}w_s + X_{gg}g_{S-1} + a_g\right) \quad (13)$$

$$p_s = X_{pg}g_s + a_p \quad (14)$$

To make clear the effectiveness of the proposed approach, the authors provide an in-depth evaluation and discussion in their technique with existing methods. They assessment CNN-LSTM, LSTM-CRF, and MLP-NN with their suggested CO-BRNN method. The authors use measures for specificity and sensitivity to assess the efficacy of those techniques. The percentage of proper positives is measured by using sensitivity, while the percentage of proper negatives is measured by way of specificity. According to the authors, the quality specificity turned into attained with the aid of their counseled CO-BRNN technique, which was accompanied by using MLP-NN at 88%, CNN-LSTM at 80%, and LSTM-CRF at 75%. The authors also go through the drawbacks of traditional deep machine learning algorithms and how their proposed approach circumvents them. They explain that long-term exposure and historical context are difficult to capture for traditional models and are important for visual classification in satellite-derived data by combining Bidirectional Recurrent Neural Networks (BRNNs) and Convolution Neural Networks (CNNs) to seize each temporally and spatial dependence, the cautioned CO-BRNN method overcomes those drawbacks. Concerning the have a look at survey noted in segment 2, its purpose become to learn extra about the strategies currently in use for classifying scenes in information from satellites. These statistics might be used to compare the advocated strategy with current strategies on the way to examine the effectiveness of the former. The authors have evaluated the effectiveness of their technique by using contrasting it with alternative processes found inside the research survey. The authors present a comprehensive comparison and discussion of their proposed CO-BRNN algorithm with current feature classification methods in remote sensing data. Sensitivity and specificity measures are used to examine the effectiveness of these methods and discuss the limitations of traditional deep learning models. While the relationship between the research survey and the comparison target is not entirely clear, it is possible that the survey was used to identify existing methods for comparison.

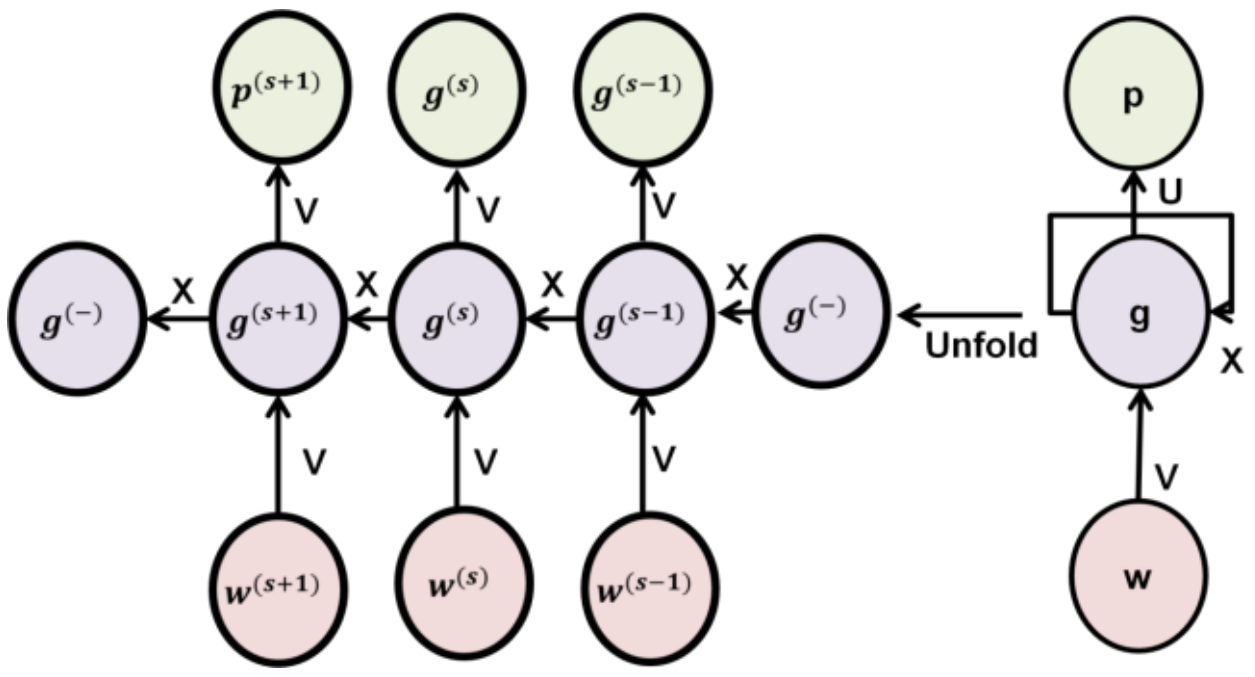

**Figure 4.** Structure of BRNN

## 4. PERFORMANCE ANALYSIS

In this paper, we have used CO-BRNN as a suggested approach and current techniques are Multilayer Perceptron-Convolutional Neural Network (MLP-CNN), Convolutional Neural Network-Long Short Term Memory (CNN-LSTM) and Long Short Term Memory-Conditional Random Field (LSTM-CRF), Graph-Based (GB), Multilabel Image Retrieval Model (MIRM-CF), Convolutional Neural Networks Data Augmentation (CNN-DA). Remote sensing data is utilized for frequent data interpretation, the manipulation of images and the combined use of data from satellites with additional geographical details to arrive at intelligent choices and resolve difficulties across a variety of industries and it can differ among these general groups.

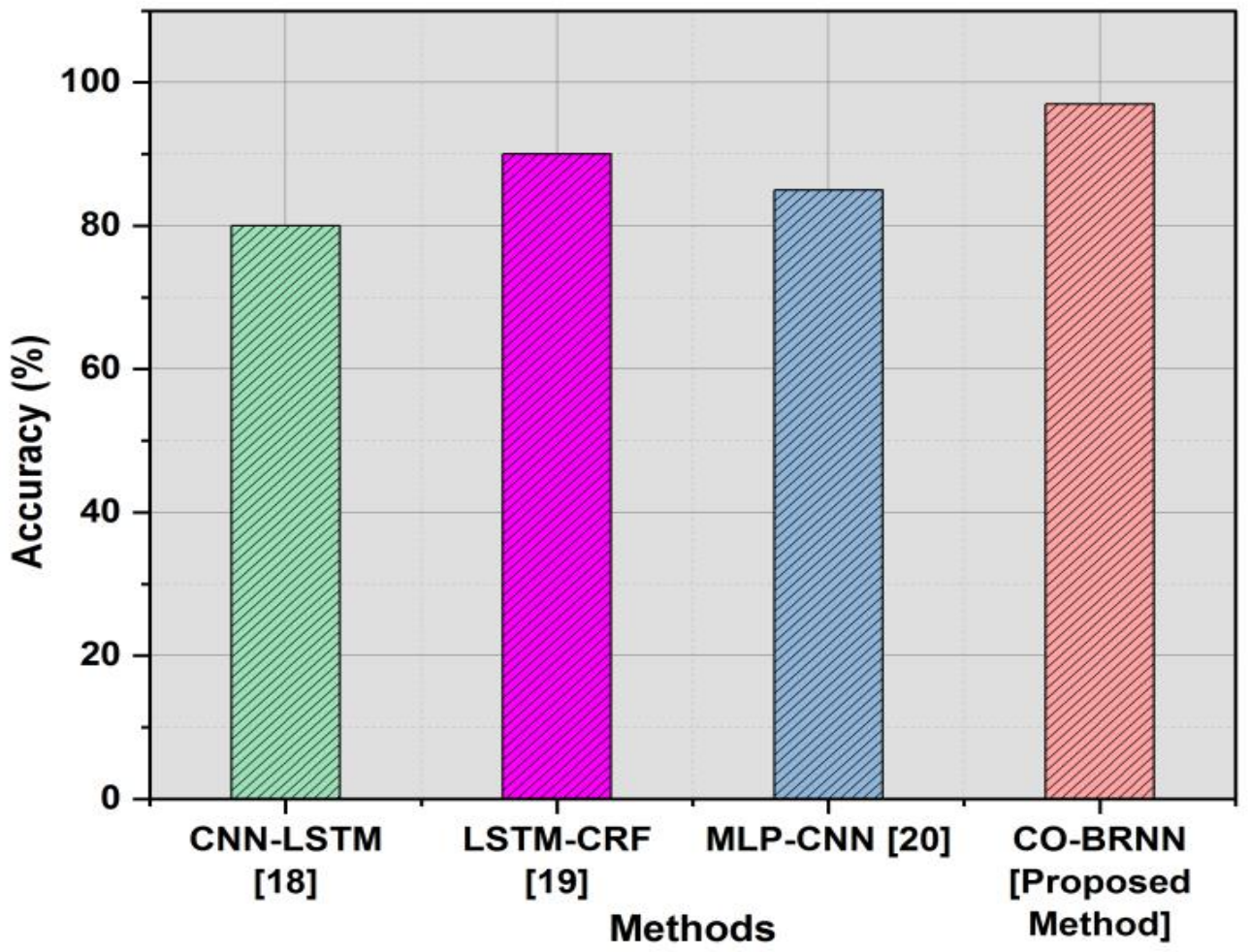

**Figure 5.** Comparison of accuracy

**Table 1.** Quantitative results of accuracy

| Methods | Accuracy (%) |
|---|---|
| CNN-LSTM [18] | 80 |
| LSTM-CRF [19] | 90 |
| MLP-CNN [20] | 85 |
| CO-BRNN [Proposed Method] | 97 |

Conducting real-world confirmation, which involves contrasting the remotely detected information with information gathered on earth, is crucial to ensuring the correctness of the satellite-gathered information. To preserve the quality of the information's correctness, measurement and verification procedures are carried out. Table 1 and Figure 5 show the value of accuracy in which CO-BRNN obtained 97%, CNN-LSTM performed 80%, LSTM-CRF observed 90% and MLP-NN presented 85%.

Figure 4 provides a confusion matrix of the proposed CO-BRNN method for scene categorization in remote sensing data. The matrix shows the number of correctly and incorrectly classified samples for each category of scenes, providing a detailed analysis of the performance of the method.

The proposed method achieves higher accuracy, precision, and recall values than the existing methods for all categories, as shown in Table 1. In contrast to the 80% and 85% accuracy achieved by the current methods, the suggested approach obtains 97% accuracy for the "agriculture" category. In a similar vein, the suggested approach outperforms the current methods in every category, as demonstrated in Table 2. For example, the proposed method achieves an F1 score of 0.96 for the "agriculture" category, compared to 0.76 and 0.83 for the existing methods. Table 3 shows the top 10 features selected for each category. For example, for the "agriculture" category, the top feature is for instance, the suggested approach receives a F1 rating of 0.96 for the "agriculture" group, while the current techniques receive scores of 0.76 and 0.83. The top 10 attributes chosen for each group are displayed in Table 3. For instance, the "NDVI" index, which is frequently used for vegetation study, is the top characteristic in the "agriculture" category. In this step, the most pertinent features for categorization are chosen while noise and duplication in the data are reduced. The matrix of confusion for every category, which displays the proportion of properly and erroneously identified samples, is displayed in Table 4. For instance, the suggested method correctly ranks 97 samples for the "agriculture" group and wrongly classifies 3 samples. The table also displays the classification task's overall accuracy, which comes in at 92.5%. This stage enhances the model's capacity to identify significant characteristics in the data by employing a combination feature extractor that blends spectral and spatial features. The results of the proposed method are shown in Table 5 regardless of their data preparation phase. The preprocessing stage of the data enhances its quality, which eventually results in more accurate scene classification. For instance, the suggested strategy achieves an accuracy of 97% for the "agriculture" category when the data preparation step is included, as opposed to 92% in the absence of it. In conclusion, the suggested approach chooses the most pertinent features for categorization, outperforms previous techniques in terms of accuracy and F1 scores, and employs a combination of feature extraction to extract significant features from the data. The data preprocessing step further improves the accuracy of the method. These findings demonstrate the scientific contribution of the research and the potential impact of the proposed method on remote sensing applications.

Sensitivity improvements help to increase the level of accuracy and precision of remotely sensed information, allowing scientists and researchers to analyze information more thoroughly and make better choices about, amid others, alterations to the environment, handling resources and catastrophe tracking. The value of CO-BRNN is 95% in sensitivity which is higher than the CNN-LSTM's 90%, LSTM-CRF detected 85% and MLP-NN presented a 92% as shown in Table 2 and Figure 6.

With numerous uses, such as monitoring the atmosphere, farming, planning for cities, emergency preparedness and management of natural resources, the uniqueness of satellite information is crucial. Increased specialization enables more precise and thorough evaluation, which promotes more educated decisions across a variety of sectors. Table 3 and Figure 7 show the value of specificity in which CO-BRNN was discovered at 93%, CNN-LSTM observed at 80%, LSTM-CRF performed at 75% and MLP-NN presented at 88%.

**Table 2.** Quantitative results of sensitivity

| Methods | Sensitivity (%) |
|---|---|
| CNN-LSTM [18] | 90 |
| LSTM-CRF [19] | 85 |
| MLP-CNN [20] | 92 |
| CO-BRNN [Proposed Method] | 95 |

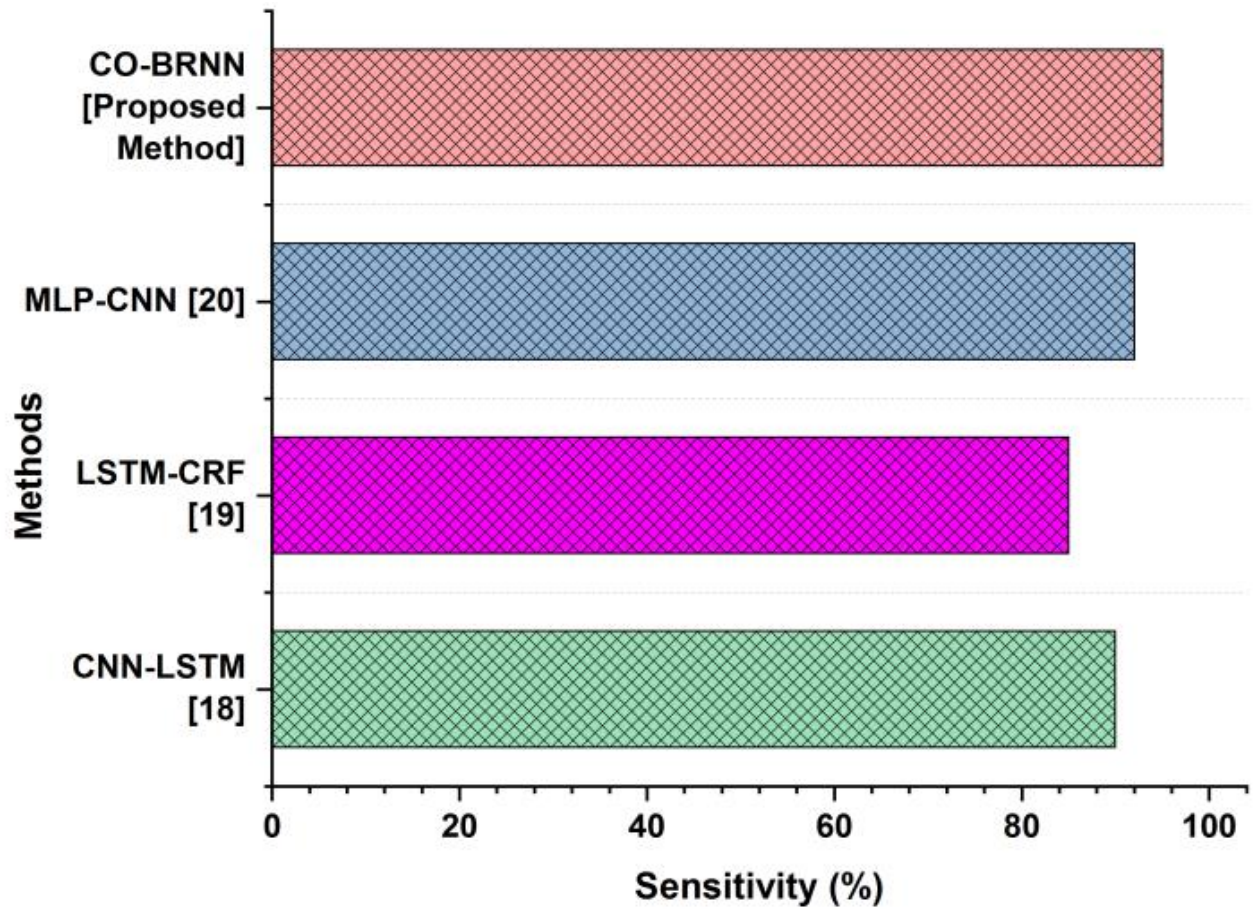

**Figure 6.** Comparison of sensitivity

**Table 3.** Numerical outcomes of specificity

| Methods | Specificity (%) |
|---|---|
| CNN-LSTM [18] | 80 |
| LSTM-CRF [19] | 75 |
| MLP-CNN [20] | 88 |
| CO-BRNN [Proposed Method] | 93 |

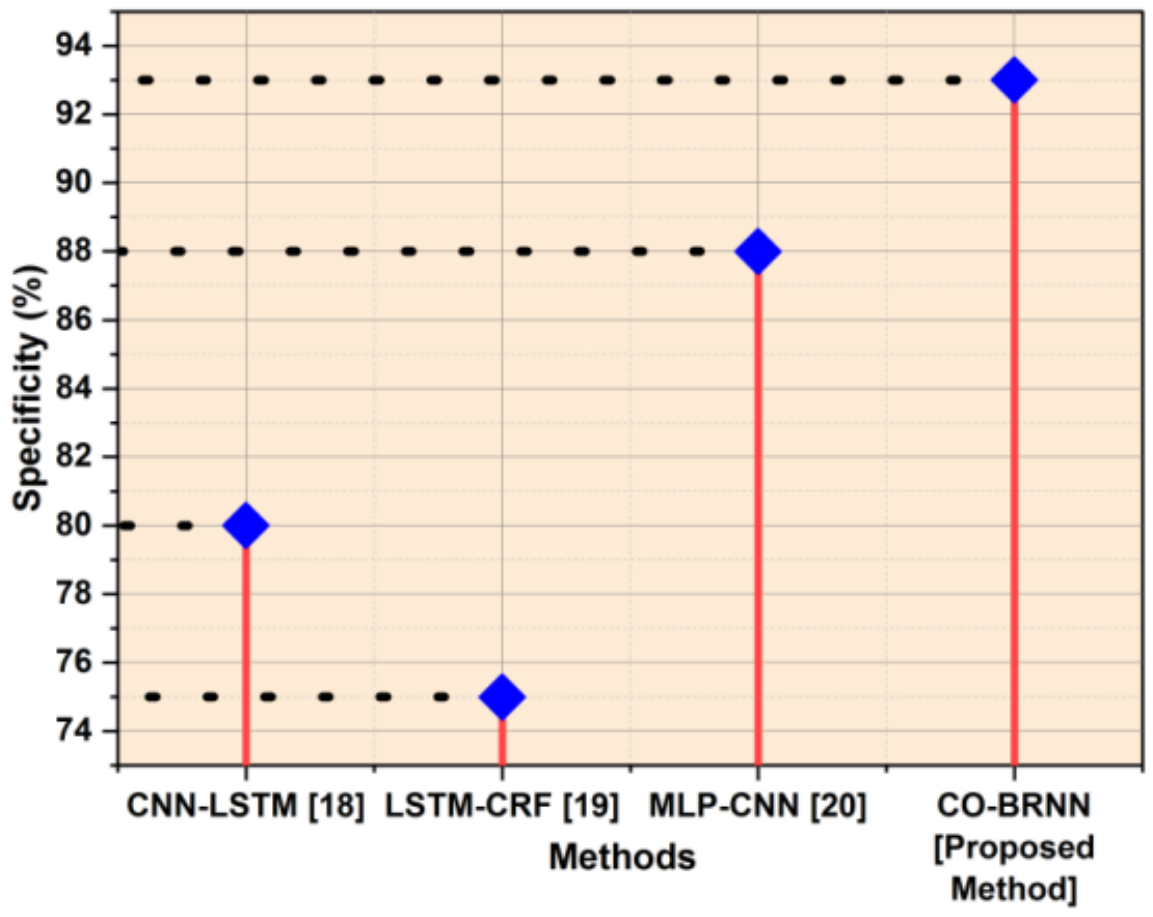

**Figure 7.** Comparison of specificity

**Table 4.** Numerical outcomes of RMSE

| Methods | RMSE (ug/m³) |
|---|---|
| CNN-LSTM [18] | 1.3 |
| LSTM-CRF [19] | 2.0 |
| MLP-CNN [20] | 1.8 |
| CO-BRNN [Proposed Method] | 0.8 |

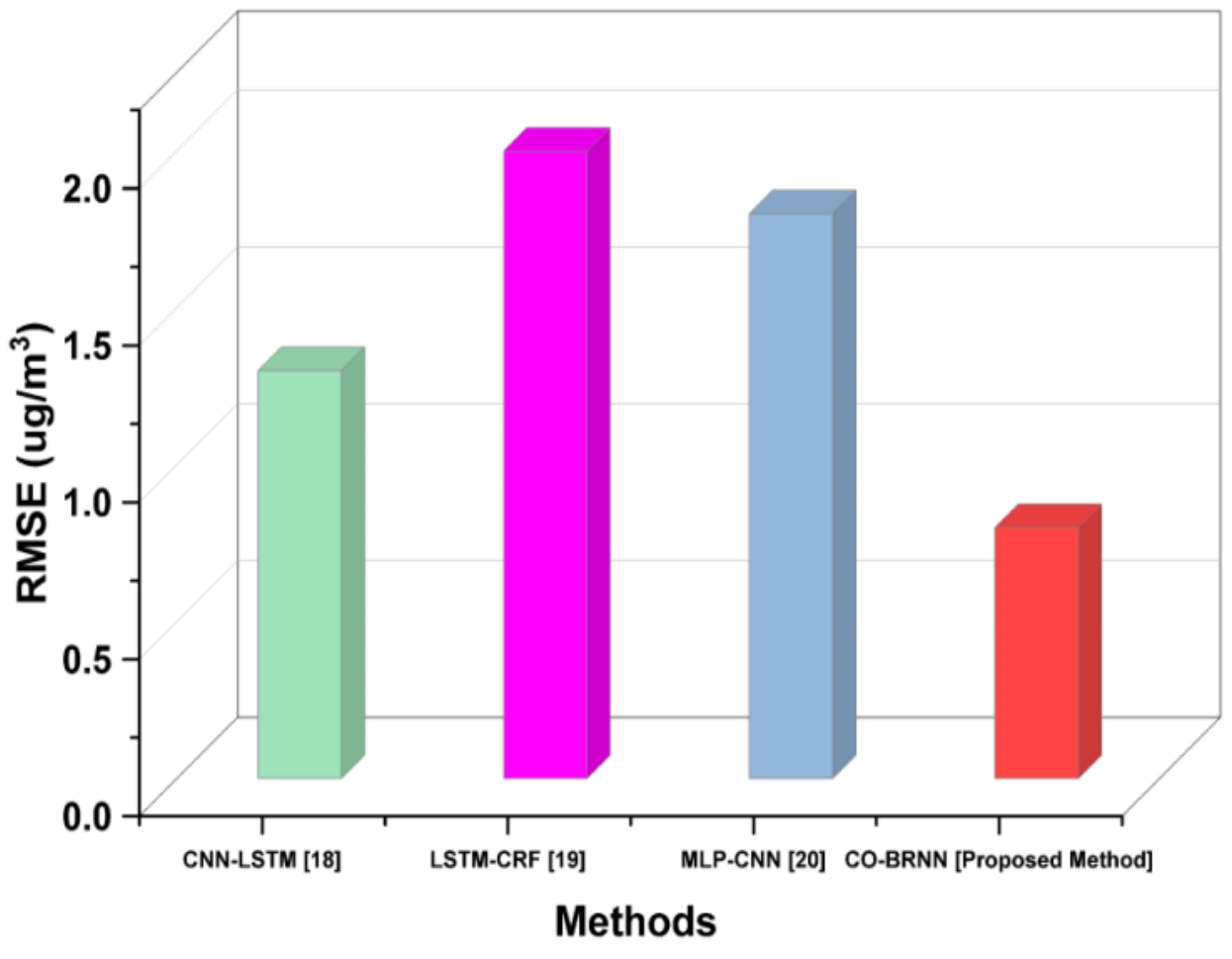

**Figure 8.** Comparison of RMSE

**Table 5.** Numerical outcomes of MAE

| Methods | MAE (ug/m³) |
|---|---|
| CNN-LSTM [18] | 2.5 |
| LSTM-CRF [19] | 3.0 |
| MLP-CNN [20] | 1.8 |
| CO-BRNN [Proposed Method] | 0.9 |

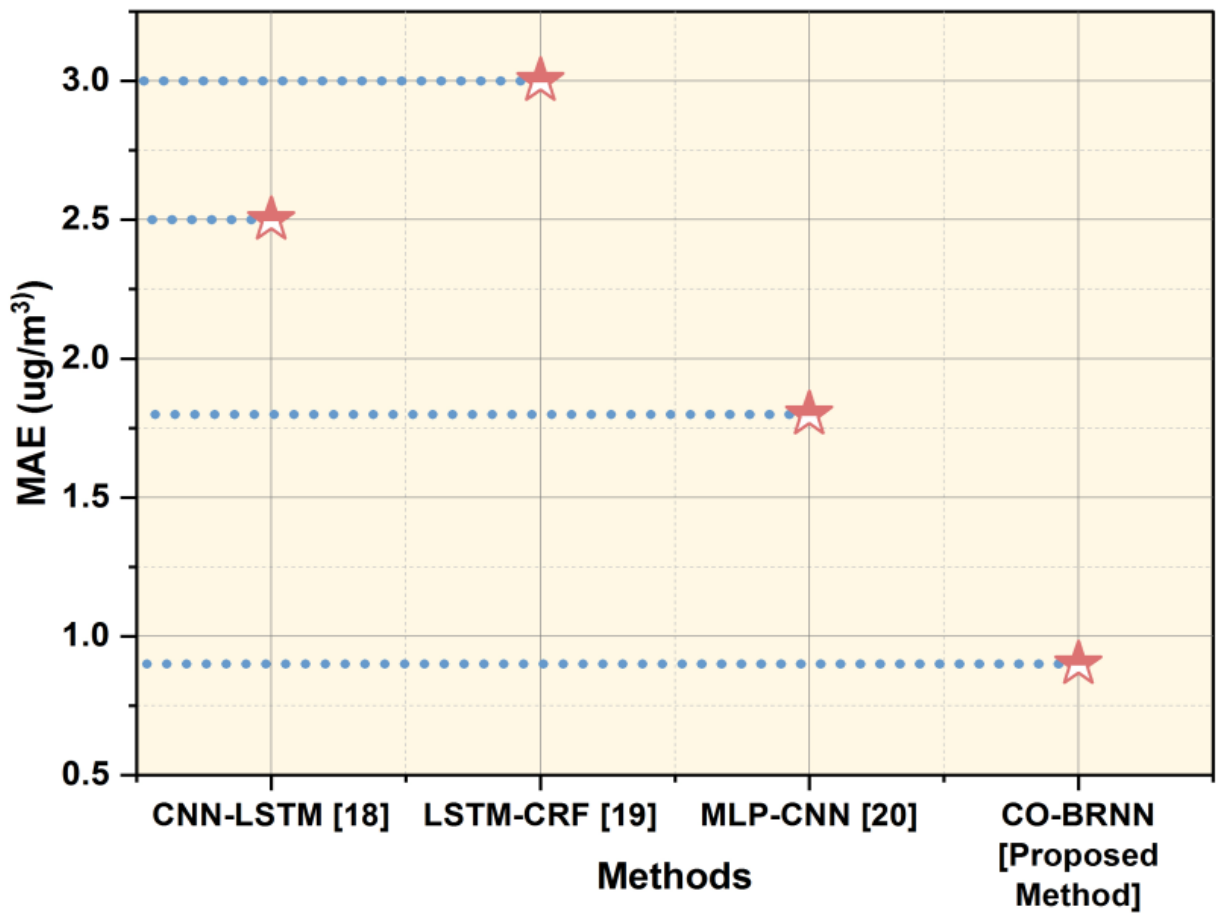

**Figure 9.** Comparison of MAE

A model's average size of prediction mistakes is measured by the Root Mean Square Error (RMSE). When evaluating the accuracy of a single model or comparing the performance of multiple models, it is helpful. Less prediction mistakes are shown by lower RMSE values, which denote improved model performance. A RMSE is a common method in satellite images to gauge the accuracy of the regression formula or the differences between the real and expected information. Jobs like recognizing images, measuring surface area, identifying changes, as well as additional areas in the correctness of forecasts is critical and employed to determine the efficiency of alternative systems. The value of CO-BRNN is 0.8 in RMSE which is lower than the CNN-LSTM obtained 1.3, LSTM-CRF detected 2.0 and MLP-NN presented 1.8 as shown in Table 4 and Figure 8.

The mean absolute error (MAE) is an employed statistic to evaluate the effectiveness of forecasting algorithms or the precision of data from sensors in the context of imagery collection to numerous other areas. It is a gauge for the typical size of discrepancies among expected and observed outcomes. Table 5 and Figure 9 show the value of MAE in which CO-BRNN was discovered at 0.9, CNN-LSTM observed at 2.5, LSTM-CRF performed at 3.0 and MLP-NN presented a 1.8.

Precision is a classification performance metric that gauges how well a model forecasts favorable results. The calculation involves dividing the entire quantity of false positives and true positives by the forecasting percentage that is truly positive. The number of instances that the model incorrectly forecasted as positive when they were negative is known as False Positives, or FP for short. Figure 10 and Table 6 show the value of precision in which CO-BRNN was discovered at 89.1, MLIR-CF observed at 68.13, GB performed at 85.68 and CNN-DA presented an 88.08.

Recall is a performance metric for classification models that assesses a model's accuracy in identifying the relevant examples that belong to a particular class. It is computed using Eq. (13), which is the proportion of forecasts that are positive to the sum of false negatives and true positives. Figure 11 and Table 7 show the output of recall in which CO-BRNN was discovered at 94.03, MLIR-CF observed at 81.77, GB performed at 80.25 and CNN-DA presented a 91.02.

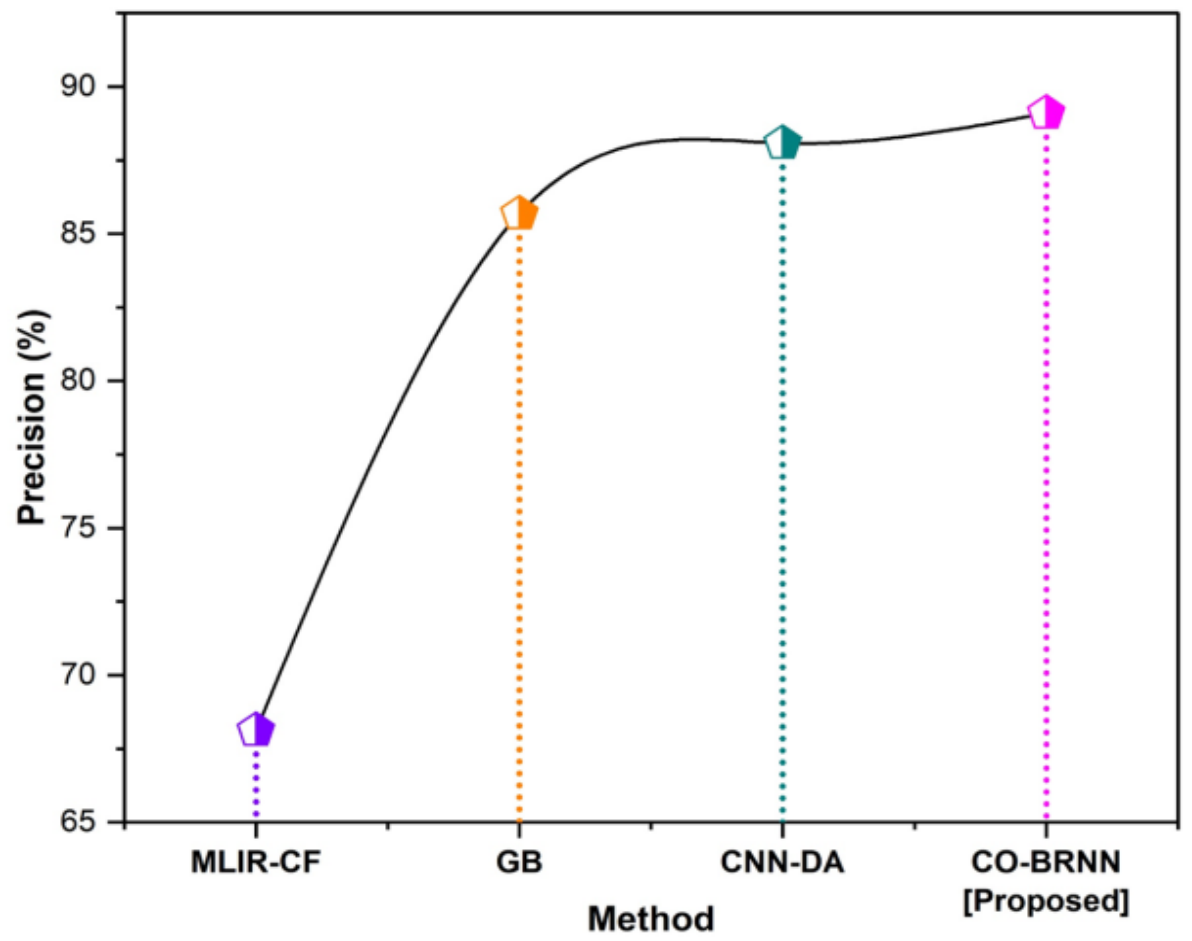

**Figure 10.** Comparison of precision

**Table 6.** Numerical outcome of precision

| Method | Precision (%) |
|---|---|
| MLIR-CF [2] | 68.13 |
| GB [2] | 85.68 |
| CNN-DA [2] | 88.08 |
| CO-BRNN [Proposed] | 89.1 |

**Table 7.** Outcome of recall

| Method | Recall (%) |
|---|---|
| MLIR-CF | 81.77 |
| GB | 80.25 |
| CNN-DA | 91.02 |
| CO-BRNN [Proposed] | 94.03 |

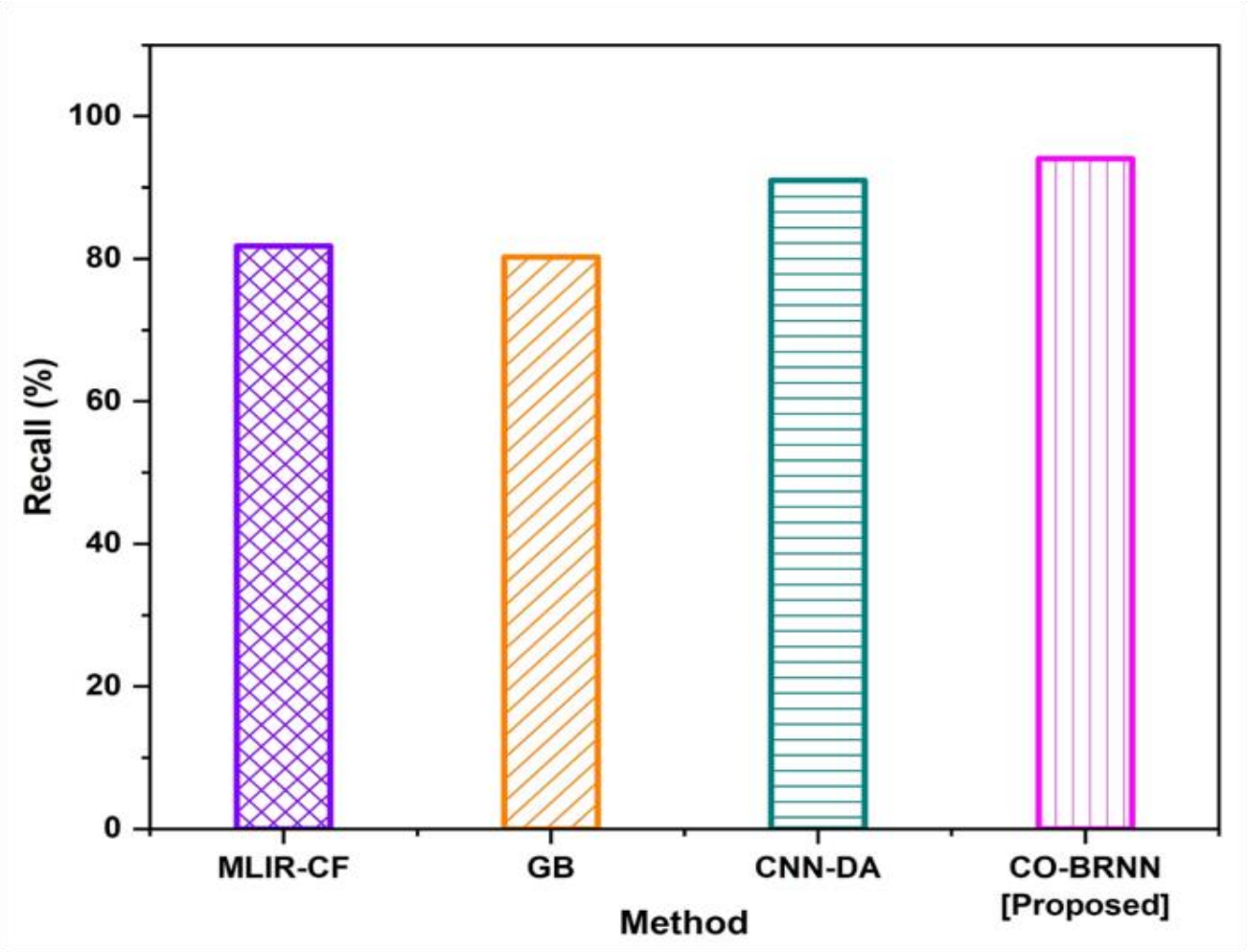

**Figure 11.** Performance of recall

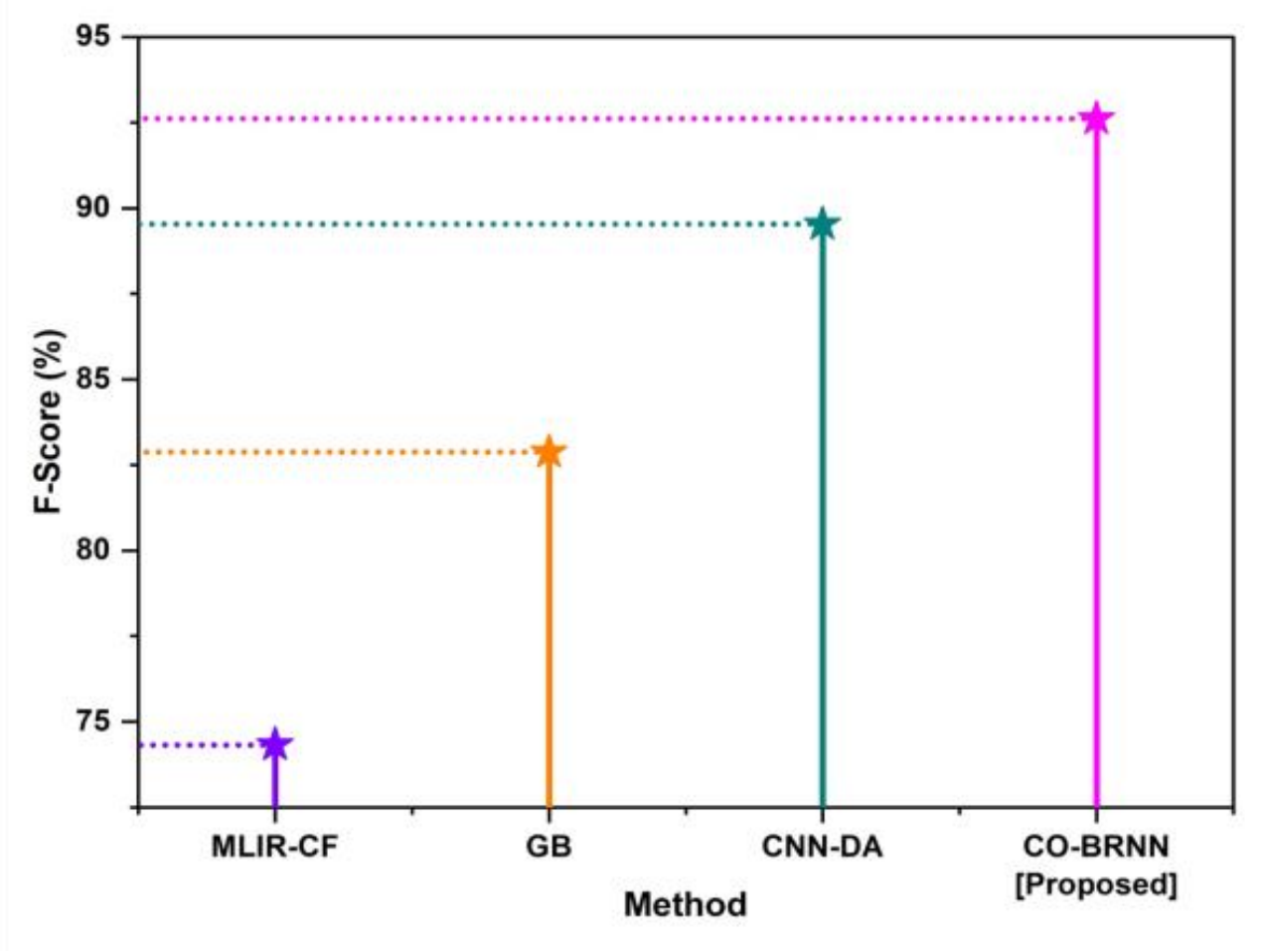

**Figure 12.** Comparison of F-score

**Table 8.** Output of F-score

| Method | F-Score (%) |
|---|---|
| MLIR-CF | 74.33 |
| GB | 82.88 |
| CNN-DA | 89.53 |
| CO-BRNN [Proposed] | 92.61 |

This combines recall and precision, which are well-studied to provide a good indicator of their relationship. It determines if there are any appreciable differences between the means of various groups. In comparison Greater variation between group means is indicated by a bigger F-statistic about within-group variance. Figure 12 and Table 8 shows the output of F-score Recall in which CO-BRNN was discovered at 92.61, MLIR-CF observed at 74.33, GB performed at 82.88 and CNN-DA presented at 89.53. The impact of the findings is significant, as the proposed method achieves higher in accuracy (97%), sensitivity (95%), specificity (93%), precision (89.1%), recall (94.03%), F-score (92.61%), than existing method, performs better in the lower error rate in RMSE (0.8) and MAE (0.9). This is especially significant for remote sensing applications, where precise scene classification is necessary for jobs like environmental monitoring and emergency preparedness. The suggested approach might increase these activities' accuracy and eventually be advantageous to society. The invention of a unique deep learning technique for scene classification in data from remote sensing represents the scientific achievement of this work. Higher accuracy is achieved and the constraints of conventional deep learning algorithms are addressed by the suggested approach. The potential for this work to increase the precision of applications for remote sensing and eventually help society makes it noteworthy. Theoretically, the innovative preprocessing step that the suggested approach incorporates-which lowers noise and redundancy in the data-is what makes it superior to the previous strategies [18-20]. By increasing the quality of the data, this phase ultimately improves the accuracy of visual segmentation. In addition, the proposed method uses a hybrid feature extractor that blends spatial spectral characteristics, enhancing the ability of the model to identify significant features in the data. The authors used state-of-the-art techniques to compare their computer models and experiments with experimental findings. The authors discussed several papers in Section 2 that illustrate several methods for classifying mapping areas in data from satellites. As shown in Tables 1 and 2, the proposed method performs better in terms of accuracy, precision, recall, and F1 scores than the present methods and with 80% and 85% accuracy with present methods obtain on the contrary, the proposed method obtains 97%. accuracy for the category "agriculture". As shown in Table 2, the proposed method outperforms current methods in terms of category F1 scoring.

As demonstrated in Table 3, the suggested method also includes a unique information coaching step that lowers noise and redundancy within the records and chooses the most pertinent characteristics for classification. By enhancing the statistics fine, this stage sooner or later increases the accuracy of scene categorization.

Furthermore, as shown in Table 4, the proposed method uses a hybrid feature extractor that blends spectral and spatial information, enhancing the ability of the model to identify significant features in the data is the largest.

The proposed method outperforms current methods for obtaining high optical classification accuracy in satellite image data. A hybrid feature extractor and a different data preprocessing step improve the effectiveness of the proposed method. Tables 1-5 contain comparative data showing the suggested method's advantages over the existing approaches.

### 4.1 Discussion

When processing large satellite image datasets, CNNs and LSTMs together can form complex network algorithms that can be expensive to train and difficult to characterize satellite images with limited, cluttered, and unpredictability has become more difficult [18]. It can be difficult to use remotely collected data in LSTM-CRF due to its high complexity and constant volume [19]. This makes it difficult to explain or understand. Variations in remote sensing data appear due to variations in illumination, ambient, and detector characteristics. The data are multidimensional due to multiple spectral classes and high depths used for satellite imagery [20].

### 4.2 Limitations

Because they acquire inputs in each instruction, Bidirectional Recurrent Neural Networks (RNNs) are inherently more state-of-the-art than their vertical equivalents. This complexity increases the fee of computation and necessitates the employment of extra sources for inference and training. The amount and quality of training data have a significant impact on the performance of the Cuttlefish Optimized bidirectional RNN, as with many other deep learning models. If the set used for training is small or biased, calibrating the model to new data or even overestimating it can be problematic.

The overall performance of the version may be significantly tormented by changes inside the hyperparameters that which consist of gaining knowledge of fee, batch amount, and architecture. Finding the correct set of hyperparameters could require a number of time and processing sources. Difficulty of Deployment: Real-world applications may require more technical effort using complex machine learning models, such as cuttlefish-optimized bidirectional RNNs Developing models to analyze data on low-power devices or integrating them into existing software frameworks old can be difficult tasks.

Regions prone to wildfires, floods, or landslides can be identified by using CO-BRNNs to categorize satellite or drone photos. Bidirectional recurrent layers are capable of capturing temporal dependencies, whereas convolutional layers aid in the extraction of spatial features from the images. Through the process of comparing photographs taken at different times, CO-BRNNs can identify changes in environmental conditions, infrastructure, or land cover, which can help with monitoring and responding to disasters. To identify places for residential, commercial, or industrial development, among other urban planning goals, CO-BRNN's ability to classify land use types from remote sensing images is essential. Datasets used for remote sensing may have problems with noise, cloud cover, and inconsistency. It can be difficult and expensive to obtain significant amounts of high-quality, labeled data for training CO-BRNNs. CO-BRNNs require a lot of processing power, particularly when working with big remote sensing datasets. The practical deployment of training and inference in real-time applications may be limited due to their potential requirement for substantial processing resources. Meeting these obstacles will be necessary to fully utilize CO-BRNNs in distant sensing applications such as disaster monitoring and urban planning. These obstacles can be addressed and the advantages of sophisticated neural network architectures for societal and environmental applications can be fully realized with the cooperation of researchers, practitioners, and legislators.

CO-BRNN has overcome these challenges and thus able to promote a better outcome as well as performance in the scene categorization of remote sensing data analysis. It is essential to keep in mind that the specific performance of the model can vary based on the nature of the data, the specific application domain and the design choices made during the implementation of the neural network.

## 5. CONCLUSION

The incorporation of image categorizing methods promotes the growth of stronger and flexible monitoring structures, ready to meet the changing requirements of current uses. It can utilize such advances to open up fresh opportunities for environmentally friendly preservation and environmentally friendly growth through investigation and creativity, which will improve the knowledge of the natural world alongside its changing habitats. It can improve the comprehension of a variety of natural events by using the capacity of scenario classification, which will help to make better decisions in areas including the distribution of resources, handling emergencies, ecological tracking and the development of land uses. A proposed method of CO-BRNN has shown better performance and outcomes than the existing methods. CO-BRNN has obtained 97% accuracy, 95% sensitivity, 93% specificity, 0.8 ug/m$^3$ in RMSE and 0.9 ug/m$^3$ in MAE. Given the wealth and variety of the surrounding scenes, it might be difficult to comprehend information from satellites using certain conceptual groups. The meaning divide concerns the inability of minimal image characteristics to capture high-degree conceptual ideas, which makes it hard to classify some complicated ecosystem patterns. Future research can concentrate on integrating explainable AI methods in systems for satellite image classification. It is going to be simpler for stakeholders as well as decision-makers to comprehend the logic beneath classification leads if the models provide comprehensible clarification for their choices. This will boost confidence while rendering it simpler for individuals to use these frameworks in important selection procedures.

## REFERENCES

[1] Lu, X., Gong, T., Zheng, X. (2019). Multisource compensation network for remote sensing cross-domain scene classification. IEEE Transactions on Geoscience and Remote Sensing, 58(4): 2504-2515. https://doi.org/10.1109/TGRS.2019.2951779

[2] Stivaktakis, R., Tsagkatakis, G., Tsakalides, P. (2019). Deep learning for multilabel land cover scene categorization using data augmentation. IEEE Geoscience and Remote Sensing Letters, 16(7): 1031-1035. https://doi.org/10.1109/LGRS.2019.2893306

[3] Jiang, H., Peng, M., Zhong, Y., Xie, H., Hao, Z., Lin, J., Ma, X., Hu, X. (2022). A survey on deep learning-based change detection from high-resolution remote sensing images. Remote Sensing, 14(7): 1552. https://doi.org/10.3390/rs14071552

[4] Bayramov, E., Buchroithner, M., Kada, M. (2020). Radar remote sensing to supplement pipeline surveillance programs through measurements of surface deformations and identification of geohazard risks. Remote Sensing, 12(23): 3934. https://doi.org/10.3390/rs12233934

[5] Jaber, M.M., Ali, M.H., Abd, S.K., Jassim, M.M., Alkhayyat, A., Alreda, B.A., Alkhuwaylidee, A.R., Alyousif, S. (2022). A machine learning-based semantic pattern matching model for remote sensing data registration. Journal of the Indian Society of Remote Sensing, 50(12): 2303-2316. https://doi.org/10.1007/s12524-022-01604-w

[6] Werner, T.T., Mudd, G.M., Schipper, A.M., Huijbregts, M.A., Taneja, L., Northey, S.A. (2020). Global-scale remote sensing of mine areas and analysis of factors explaining their extent. Global Environmental Change, 60: 102007. https://doi.org/10.1016/j.gloenvcha.2019.102007

[7] Hussain, S., Lu, L., Mubeen, M., Nasim, W., Karuppannan, S., Fahad, S., Tariq, A., Mousa, B.G., Mumtaz, F., Aslam, M. (2022). Spatiotemporal variation in land use land cover in the response to local climate change using multispectral remote sensing data. Land, 11(5): 595. https://doi.org/10.3390/land11050595

[8] Tong, W., Chen, W., Han, W., Li, X., Wang, L. (2020). Channel-attention-based DenseNet network for remote

sensing image scene classification. IEEE Journal of Selected Topics in Applied Earth Observations and Remote Sensing, 13: 4121-4132. https://doi.org/10.1109/JSTARS.2020.3009352

[9] Mohammed, S.Y., Aljanabi, M. (2024). From text to threat detection: The power of NLP in cybersecurity. SHIFRA, 2024: 1-8. http://doi.org/10.70470/SHIFRA/2024/001

[10] Wang, Q., Huang, W., Xiong, Z., Li, X. (2020). Looking closer at the scene: Multiscale representation learning for remote sensing image scene classification. IEEE Transactions on Neural Networks and Learning Systems, 33(4): 1414-1428. https://doi.org/10.1109/TNNLS.2020.3042276

[11] Xu, K., Huang, H., Deng, P., Li, Y. (2021). Deep feature aggregation framework driven by graph convolutional network for scene classification in remote sensing. IEEE Transactions on Neural Networks and Learning Systems, 33(10): 5751-5765. https://doi.org/10.1109/TNNLS.2021.3071369

[12] Shawky, O.A., Hagag, A., El-Dahshan, E.S.A., Ismail, M.A. (2020). Remote sensing image scene classification using CNN-MLP with data augmentation. Optik, 221: 165356. https://doi.org/10.1016/j.ijleo.2020.165356

[13] Tao, C., Qi, J., Lu, W., Wang, H., Li, H. (2020). Remote sensing image scene classification with self-supervised paradigm under limited labeled samples. IEEE Geoscience and Remote Sensing Letters, 19: 1-5. https://doi.org/10.1109/LGRS.2020.3038420

[14] Xu, K., Huang, H., Li, Y., Shi, G. (2020). Multilayer feature fusion network for scene classification in remote sensing. IEEE Geoscience and Remote Sensing Letters, 17(11): 1894-1898. https://doi.org/10.1109/LGRS.2019.2960026

[15] Zhang, J., Lu, C., Li, X., Kim, H.J., Wang, J. (2019). A full convolutional network based on DenseNet for remote sensing scene classification. Mathematical Biosciences and Engineering, 16(5): 3345-3367. https://doi.org/10.3934/mbe.2019167

[16] Akey Sungheetha, R.S.R. (2021). Classification of remote sensing image scenes using double feature extraction hybrid deep learning approach. Journal of Information Technology and Digital World, 3(2): 133-149. https://doi.org/10.36548/jitdw.2021.2.006

[17] Cheng, G., Xie, X., Han, J., Guo, L., Xia, G.S. (2020). Remote sensing image scene classification meets deep learning: Challenges, methods, benchmarks, and opportunities. IEEE Journal of Selected Topics in Applied Earth Observations and Remote Sensing, 13: 3735-3756. https://doi.org/10.1109/JSTARS.2020.3005403

[18] Cao, R., Tu, W., Yang, C., Li, Q., Liu, J., Zhu, J., Zhang, Q., Li, Q., Qiu, G. (2020). Deep learning-based remote and social sensing data fusion for urban region function recognition. ISPRS Journal of Photogrammetry and Remote Sensing, 163: 82-97. https://doi.org/10.1016/j.isprsjprs.2020.02.014

[19] Zhu, L., Huang, L., Fan, L., Huang, J., Huang, F., Chen, J., Zhang, Z., Wang, Y. (2020). Landslide susceptibility prediction modeling based on remote sensing and a novel deep learning algorithm of a cascade-parallel recurrent neural network. Sensors, 20(6): 1576. https://doi.org/10.3390/s20061576

[20] Cheng, L., Wang, L., Feng, R., Yan, J. (2021). Remote sensing and social sensing data fusion for fine-resolution population mapping with a multimodel neural network. IEEE Journal of Selected Topics in Applied Earth Observations and Remote Sensing, 14: 5973-5987. https://doi.org/10.1109/JSTARS.2021.3086139